\documentclass{article} 
\usepackage{iclr2025_conference,times}

\usepackage{amsmath,amsfonts,bm}

\def\eqref#1{equation~\ref{#1}}

\def\1{\bm{1}}

\DeclareMathAlphabet{\mathsfit}{\encodingdefault}{\sfdefault}{m}{sl}
\SetMathAlphabet{\mathsfit}{bold}{\encodingdefault}{\sfdefault}{bx}{n}

\usepackage{hyperref}
\usepackage{url}
\usepackage{graphicx}
\usepackage{subcaption}
\usepackage{stmaryrd}
\usepackage{svg}
\usepackage{amsthm}
\usepackage{amssymb}
\usepackage{wrapfig}

\title{	
Value-Guided Action Planning with JEPA World Models}

\iclrfinalcopy

\author{
Matthieu Destrade $^{1,2}$\thanks{Correspondence: \texttt{matthieu.destrade@polytechnique.edu}} ,
Oumayma Bounou$^{3}$,
Quentin Le Lidec$^{3}$,
Jean Ponce$^{2,3}$,
Yann LeCun$^{3}$ \\
$^{1}$ École Polytechnique \; $^{2}$ ENS Paris \; $^{3}$ New York University
}

\begin{document}

\maketitle
\vspace{-0.5cm}
\begin{abstract}
Building deep learning models that can reason about their environment requires capturing its underlying dynamics. Joint-Embedded Predictive Architectures (JEPA) provide a promising framework to model such dynamics by learning representations and predictors through a self-supervised prediction objective. However, their ability to support effective action planning remains limited. We propose an approach to enhance planning with JEPA world models by shaping their representation space so that the negative goal-conditioned value function for a reaching cost in a given environment is approximated by a distance (or quasi-distance) between state embeddings. We introduce a practical method to enforce this constraint during training and show that it leads to significantly improved planning performance compared to standard JEPA models on simple control tasks.
\end{abstract}

\section{Introduction}
World models are a class of deep learning architectures designed to capture the dynamics of systems (\cite{WM, ding2025understandingworldpredictingfuture}). They are trained to predict future states of an environment given a sequence of actions. By explicitly modeling the system’s dynamics, they capture a causal understanding of how actions influence future outcomes, enabling reasoning and planning over possible trajectories.

Among the various architectures proposed to implement such models, Joint-Embedded Predictive Architectures (JEPA) (\cite{LeCun2022APT}) provide an effective framework for learning predictive representations. By optimizing a self-supervised prediction loss, JEPA models jointly learn representations of states and predictors that map past states and actions to future representations. This formulation has proven effective for both representation learning (\cite{assran2023selfsupervisedlearningimagesjointembedding,bardes2024revisitingfeaturepredictionlearning}) and action planning (\cite{sobal2025learningrewardfreeofflinedata,zhou2025dinowmworldmodelspretrained}), the latter referring to the optimization of action sequences that drive a system from an initial state to a goal state.

In this work, we aim to enhance the planning capabilities of JEPA models. Inspired by advances in reinforcement learning, we learn representations such that the Euclidean distance (or a quasi-distance) between embedded states approximates the negative goal-conditioned value function associated with a reaching cost (\cite{hilp,park2024hiqlofflinegoalconditionedrl,wang2023optimalgoalreachingreinforcementlearning}). This structure provides a meaningful latent representation space for planning, potentially mitigating local minima during planning optimization. We evaluate our method on control tasks and observe that incorporating such representations consistently improves planning performance compared to standard JEPA models.

\section{Related work}
\subsection{JEPA world models}
 Joint-Embedded Predictive Architectures (JEPA) (\cite{LeCun2022APT}) provide an effective way to implement world models for representation learning and action planning. They rely on the hypothesis that predicting future states is easier in a learned representation space than in the original observation space, and that enforcing predictability encourages meaningful representations. A JEPA model typically consists of a state encoder, an action encoder, and a predictor. It is trained on sequences of states and actions by minimizing a prediction loss, $\mathcal{L}_\text{pred}$, between a predicted representation and that of the actual state resulting from applying a given action. To prevent collapse during training, standard approaches use a VCReg loss, $\mathcal{L}_\text{VCReg}$, as in \cite{sobal2025learningrewardfreeofflinedata}, or an exponential moving average (EMA) scheme, as in \cite{assran2023selfsupervisedlearningimagesjointembedding, bardes2024revisitingfeaturepredictionlearning}.
 
Recent works (\cite{sobal2025learningrewardfreeofflinedata, zhou2025dinowmworldmodelspretrained}) have applied JEPA models to action-planning tasks, showing promising yet still limited performance. To do so, they employ a model predictive control (MPC) procedure (\cite{GARCIA1989335}), which iteratively minimizes a planning loss measuring the distance between predicted and goal representations over a finite horizon.

\subsection{Learning a value function}

To improve the effectiveness of MPC, several works have proposed learning a value function to guide the MPC procedure \cite{farshidian2019deepvaluemodelpredictive, jordana2025infinitehorizonvaluefunctionapproximation}. This approach allows MPC to account for longer time horizons, and can stabilize the procedure by providing an additional cost term whose minimization facilitates goal-reaching tasks.

Implicit Q-Learning (IQL) \cite{ghosh2023reinforcementlearningpassivedata, kostrikov2021offlinereinforcementlearningimplicit,xu2023policyguidedimitationapproachoffline} learns a goal-conditioned value function from unlabeled trajectories by leveraging expectile regression. The authors of \cite{hilp} leverage IQL to learn a structured representation space for states of a system, where the negative Euclidean distance approximates a goal-conditioned value function corresponding to the terminal cost in a reaching objective. They show that these representations enable solving various reinforcement learning tasks efficiently. Since a goal-conditioned value function is not symmetric in general, additional work has proposed learning it using a quasi-distance \cite{wang2023optimalgoalreachingreinforcementlearning}.

\section{Value-guided JEPA for action planning}

To improve the planning capabilities of JEPA models, we focus on enhancing the representations used to compute the MPC planning cost. In the standard JEPA framework, planning is performed by minimizing the distance between a predicted state and the goal in the representation space. However, this cost can have numerous local minima, making optimization challenging. To address this, we propose learning representations such that the Euclidean distance in the representation space corresponds to the negative of the goal-conditioned value function associated with a reaching cost in a given environment, as in \cite{hilp}. Unlike previous works, we focus on using these representations for planning with JEPA models and MPC procedures, rather than solely for policy execution. Under this formulation, setting the planning cost to the learned value function and minimizing it naturally drives the model toward the goal.

\subsection{Baseline loss functions}
To enforce the value function criterion in the representation space, we consider several simple loss functions for the state encoder of a JEPA model, which serve as baselines. Specifically, we apply a contrastive loss $\mathcal{L}_\text{contrastive}$ using successive states from training trajectories as positive examples and random pairs of states as negative examples, as well as a regression loss $\mathcal{L}_\text{regressive}$ explicitly enforcing the distance between successive states to be 1.

\subsection{IQL for JEPA models}
Let $\mathcal{S}_0$ be the state space, $\theta$ the parameters and $\mathcal{E}_\theta$ the state encoder of a JEPA model. For all \mbox{$(s, g) \in \mathcal{S}_0^2$}, we define $V_\theta(s, g) = -\Vert \mathcal{E}_\theta(s) - \mathcal{E}_\theta(g) \Vert_2$. Our goal is to learn $\theta$ such that $V_\theta$ approximates the goal-conditioned value function $V^\star$ associated with the reaching cost \mbox{$C: (s, a, g) \mapsto \mathbf{1}_{s \neq g}$}, which penalizes all time steps where the state $s$ is not equal to the goal $g$.

Let $(T, N) \in \mathbb{N}^2$ represent the length of the training trajectories and the number of training goals. Let $\mathcal{D}$ be a dataset of trajectories $(s_t)_{t\in\llbracket0,T\rrbracket}$ belonging to $\mathcal{S}_0^{T+1}$ and goals $(g_n)_{n\in\llbracket0,N\rrbracket}$  belonging to $\mathcal{S}_0^{N+1}$. We minimize the mean IQL loss with respect to $\theta$ via gradient descent:
\begin{equation}
\forall ((s_t),(g_n)) \in \mathcal{D}, \quad
\mathcal{L}_\text{VF}^\theta((s_t),(g_n)) = \sum_{n=0}^N \sum_{t=0}^{T-1} 
L_\tau^2 \Big( -\mathbf{1}_{s_t \neq g_n} + \gamma V_{\bar{\theta}}(s_{t+1}, g_n) - V_\theta(s_t, g_n) \Big),
\end{equation}
where $\bar{\cdot}$ denotes a stop-gradient ; $\tau, \gamma \in ]0,1[$ are close to $1$ ; and for all $x \in \mathbb{R},$ the term \mbox{$\; L_\tau^2(x) = |\tau - \mathbf{1}_{x<0}| \, x^2$} performs expectile regression. The parameter $\gamma$ is the discount factor of the value function we aim to learn.
In practice, we use two different types of goals: the last state of the training trajectories, and random goals sampled from the training batches.

To obtain a better approximation, we further explore replacing the Euclidean distance in the definition of $V_\theta$ with a quasimetric distance, following \cite{wang2023optimalgoalreachingreinforcementlearning}. The quasi-distance used to learn $V^\star$ is the generic form introduced in \cite{wang2022improved}.

We consider two approaches to training JEPA models. The first approach, which we call ``Sep'', consists of training the state encoder alone using the $\mathcal{L}_\text{VF}$ objective, followed by training the action encoder and predictor with the $\mathcal{L}_\text{pred}$ loss. The second approach consists of training all networks together using as objective the sum of $\mathcal{L}_\text{VF}$ and $\mathcal{L}_\text{pred}$.

\section{Experiments}

\subsection{Experiment settings}

We conduct our experiments in two environments under an offline reinforcement learning setting. Models are trained with random trajectories sampled in the environments. The states used as inputs to our models are observation images, potentially including additional sensory information. A detailed description of the datasets used is provided in the Appendix \ref{app:dataset}.

The wall environment consists of a square space separated by a wall with a door. The positions of the wall and door are randomly initialized when the environment is instantiated. The agent has to move from a random starting position to a random goal located on the opposite side of the wall. It can execute actions that are vectors corresponding to displacements. We generate datasets with two settings: WS, with actions of small norms, and WB, with actions of larger norms.

The maze environment consists of an agent that must move from a random starting point to a random goal within a random maze. Its actions are velocity commands. Planning in this environment requires that both the agent's position and velocity be encoded in the representations, as it simulates inertia. Following a similar approach to \cite{sobal2025learningrewardfreeofflinedata}, we include the agent's velocity as an input to the encoders for a given state.

\subsection{Planning with the representations}
We conduct experiments to evaluate the planning performance of different learning methods. Specifically, we train JEPA models with:

  \begin{table}[h]
  \centering
  
  \begin{minipage}{0.47\textwidth}
    \centering
    \begin{tabular}{|c||c|c|}
        \hline
        Name& State encoder loss&Sep\\
        \hline\hline
        Contrastive & $\mathcal{L}_\text{contrastive}$ & \checkmark\\
        \hline
        Regressive & $\mathcal{L}_\text{regressive}$ \& $\mathcal{L}_\text{VCReg}$ & \checkmark\\
        \hline
        pred\_VCReg& $\mathcal{L}_\text{VCReg}$ &$\times$\\
        \hline
        pred\_EMA& EMA procedure &$\times$\\
        \hline
        VF & $\mathcal{L}_\text{VF}$ & \checkmark \\
        \hline
    \end{tabular}
  \end{minipage}
  \hfill
  \begin{minipage}{0.52\textwidth}
    \centering
    \begin{tabular}{|c||c|c|}
        \hline
        Name& State encoder loss &Sep\\
        \hline\hline
        VF\_pred & $\mathcal{L}_\text{VF}$  & $\times$ \\
        \hline
        VF\_quasi &$\mathcal{L}_\text{VF}$ \& quasi-distance& \checkmark\\
        \hline
        VF\_quasi\_pred & $\mathcal{L}_\text{VF}$ \& quasi-distance& $\times$\\
        \hline
        VF\_VCReg & $\mathcal{L}_\text{VF}$ \& $\mathcal{L}_\text{VCReg}$& \checkmark\\
        \hline
        VF\_VCReg\_pred& $\mathcal{L}_\text{VF}$ \& $\mathcal{L}_\text{VCReg}$& $\times$\\
        \hline
    \end{tabular}
  \end{minipage}

  \caption{Training approaches}
  \label{tab:exemple}
\end{table}

The precise settings of the experiments are described in Appendix \ref{Expsettings}.

We assess the quality of the learned representations by evaluating the planning accuracy of the model, defined as the proportion of successful plans for random pairs of initial states and goals. We compute this success rate on 200 instances of the wall environment and 80 of the maze one, so that the variance of the results is small. We use an MPC procedure with an MPPI optimizer. The results are displayed in Table \ref{tab:results}.

  \begin{table}[h]
  \centering
  
  \begin{minipage}{0.49\textwidth}
    \centering
    \begin{tabular}{|c||c|c|c|}
        \hline
        Type & WS & WB & Maze\\
        \hline \hline
        Contrastive & 0.49& 0.59&0.50\\
        \hline
        Regressive & 0.54&0.57&0.46\\
        \hline
        pred\_VCReg & 0.55&0.89&0.54\\
        \hline
        pred\_EMA & 0.46&0.43&0.04\\
        \hline
        VF & 0.63&0.94&0.49\\
        \hline
    \end{tabular}
  \end{minipage}
  \hfill
  \begin{minipage}{0.49\textwidth}
    \centering
    \begin{tabular}{|c||c|c|c|}
        \hline
        Type & WS & WB & Maze\\
        \hline \hline
        VF\_pred & 0.55  & 0.75&0.49\\
        \hline
        VF\_quasi &\textbf{ 0.71 }& \textbf{0.96}&\textbf{0.63}\\
        \hline
        VF\_quasi\_pred & 0.61 & 0.85&0.43\\
        \hline
        VF\_VCReg & 0.49 &0.75&0.39\\
        \hline
        VF\_VCReg\_pred & 0.47&0.67&0.39\\
        \hline
    \end{tabular}
  \end{minipage}

  \caption{Planning results in the different environments}
  \label{tab:results}
\end{table}

They show that IQL-inspired approaches provide valuable guidance during planning and achieve better results than intuitive or prediction-based approaches, as used in \cite{sobal2025learningrewardfreeofflinedata}. Interestingly, the VF\_quasi approach consistently outperforms the VF approach, even when the theoretical value function is symmetric. This suggests that using a quasi-distance always facilitates the training process by enhancing the expressiveness of the networks.

Learning representations using both a prediction loss and an IQL loss is less effective than using the latter loss alone. Using VCReg to promote diversity when learning with an IQL loss also results in poor planning performance. The results obtained with the WB dataset are better than those obtained with the WS dataset. This may be due to the fact that a single trajectory explores more of the environment in the WB dataset, and that the agent is more likely to collide with the wall.

\section{Discussion}

\noindent \textbf{Locality of the training.} The imperfect results indicate that the value functions learned with our approach are inaccurate. While local relationships between states can be expected to be correctly captured, this is less probable for distant relationships, for two main reasons. First, during training, the space of distant triplets of states (starting state, following state and goal) is sparsely sampled. Second, the gradient of the discounted value function with respect to the state becomes small when the state is far from the given goal. For such states, the signal-to-noise ratio of the value function tends to be low. This suggests that using a hierarchy of representation spaces, where higher levels model longer-range transitions or more coarsely sampled trajectories, may better capture distant relationships and yield improved results.

\noindent \textbf{Influence of the dataset.} Theoretical results on the IQL loss show that only the support of the policy used to create the training dataset actually matters when $\tau$ tends to $1$. In practice, however, other factors are likely relevant. In highly suboptimal trajectories, states that are close to each other may appear far apart, potentially making training more difficult. Therefore, it might be preferable to use ``expert'' trajectories. However, they are often hard to obtain and come at the cost of diversity and exploration. Moreover, it is important that the states used in the IQL loss during training span the entire state space. In practice, this can be achieved either by increasing the size of the training dataset or by employing more effective data collection strategies that better explore underrepresented states.


\section{Conclusion}

In this study, we aimed to improve the planning capabilities of JEPA world models. To this end, we proposed enhancing the representations used for planning by learning them such that the Euclidean distance, or a quasi-distance, in the representation space approximates the negative goal-conditioned value function associated with a goal-reaching cost for the system under consideration. This was achieved by training the state encoder of a JEPA model using an implicit Q-learning (IQL) loss.

We compared these methods to more intuitive approaches, as well as to standard prediction-based JEPA training approaches, on benchmark action-planning tasks. Our results show that the value function–based methods, particularly those using a quasi-distance, achieve superior performance, suggesting that such approaches are a promising direction for world model action planning.

Further experiments would be valuable, especially in random environments. Prediction-based methods are indeed expected to be more robust to stochasticity in non-deterministic environments and may enable the learning of more general representations than the other approaches tested, whereas our IQL approach is known to be biased in random environments.

\newpage
\bibliography{iclr2025_conference}
\bibliographystyle{iclr2025_conference}

\newpage
\section{Appendix}

\subsection{Datasets}
\label{app:dataset}
 
\subsubsection{Wall}

\begin{wrapfigure}{r}{0.58\textwidth} 
    \centering
    \begin{minipage}[b]{0.29\textwidth}
        \centering
        \includegraphics[width=\linewidth]{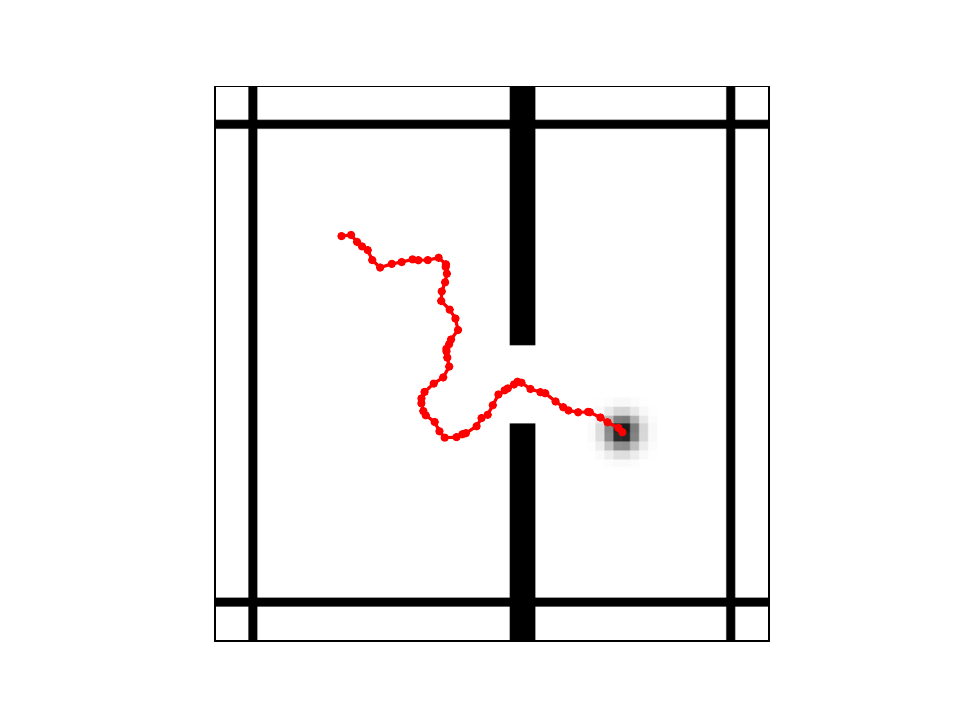}
        \caption*{Crossing trajectory, WS}
    \end{minipage}\hfill
    \begin{minipage}[b]{0.29\textwidth}
        \centering
        \includegraphics[width=\linewidth]{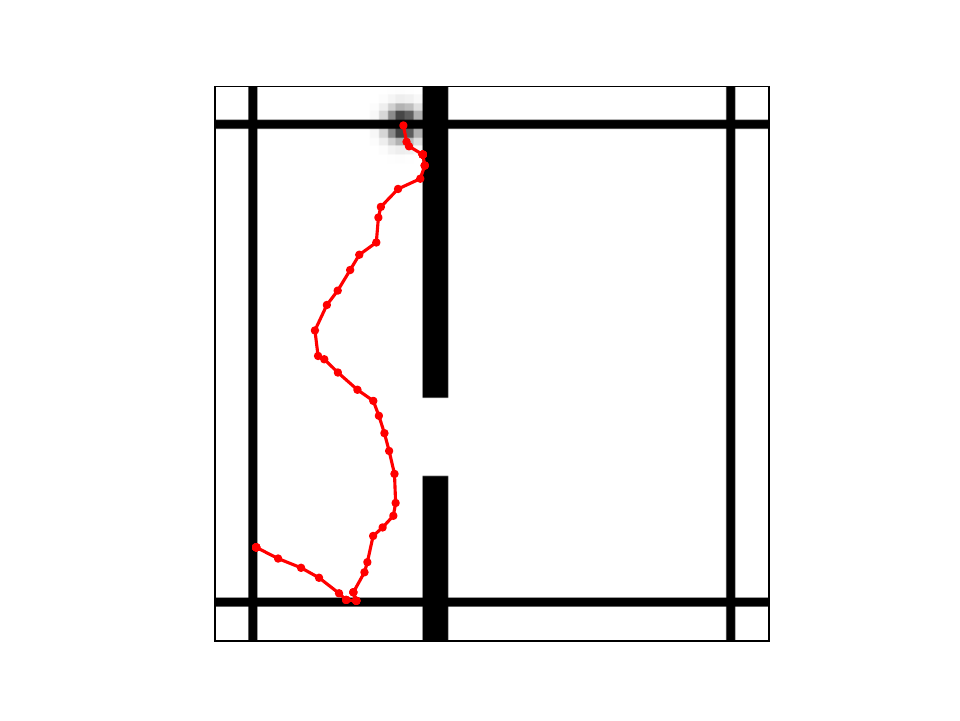}
        \caption*{Non-crossing trajectory, WB}
    \end{minipage}
    \caption{Examples of trajectories from the wall datasets}
    \label{fig:wallsmall}
\end{wrapfigure}

The observations of states in the wall environment are of size $64\times64$ and consist of 2 channels: one representing the agent and the other representing the walls. Visualizations of typical states of this environment (with flattened channels) are shown in Fig. \ref{fig:wallsmall}.

To generate the dataset of training trajectories, we follow the approach of \cite{sobal2025learningrewardfreeofflinedata}, and do not sample actions using Gaussian noise, as this would result in trajectories concentrated in a small region of the environment. Instead, we generate actions by sampling a random direction, perturbing it with noise drawn from the von Mises distribution with concentration parameter 5. We generate datasets containing 1000 trajectories of length 64, ensuring that half of the trajectories correspond to the agent passing through the door.

The WS dataset is generated with action norms sampled randomly from a Gaussian distribution with mean 1 pixel and standard deviation 0.4, and clipped to the range $[0.2,1.8]$. The WB dataset is generated with action norms sampled randomly from a Gaussian distribution with mean 2 pixels and standard deviation 0.8, and clipped to the range $[0.4,3.6]$.

\subsubsection{Maze}

The maze environment follows the setting used in \cite{sobal2025learningrewardfreeofflinedata}, which is based on the Mujoco PointMaze environment \cite{fu2021d4rldatasetsdeepdatadriven}. It uses a grid of $4\times4$ squares, of which between $50\%$ and $60\%$ contiguous squares are selected to form the maze. Observations of states in this environment are of size $64\times64$, are colored, and have 3 channels. 

The actions controlling the agent correspond to target speeds to reach. The environment computes the force required to achieve the desired speed after a certain number of time steps. The trajectories are generated by sampling random speed vectors with norms smaller than 5, starting from random positions. To evaluate the planning capabilities of our approaches in this environment, random starting points and goals are sampled, such that they are at least 3 cells apart. The dataset contains 1000 trajectories of length 101.

To assess the generalization capabilities of the different approaches we experiment with, we follow the methodology of \cite{sobal2025learningrewardfreeofflinedata}. The training trajectories all belong to five maze layouts, that are different from those used for evaluation. 

\begin{figure}[h]
    \centering
    \begin{subfigure}[b]{0.18\textwidth}
        \centering
    \includegraphics[width=1.0\linewidth]{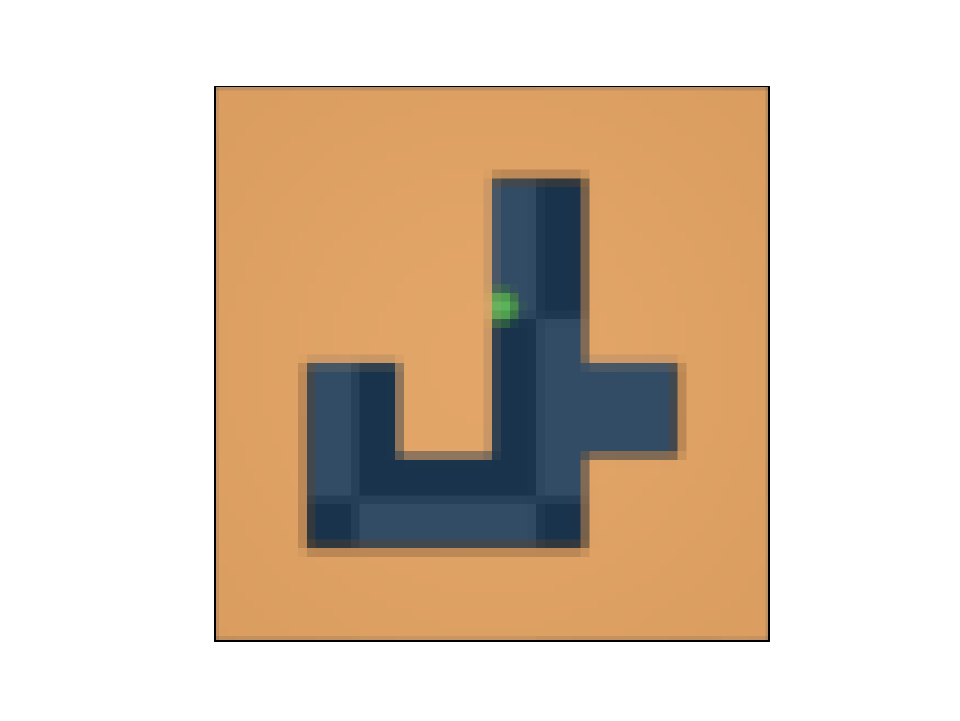} 
    \end{subfigure}%
    ~
    \begin{subfigure}[b]{0.18\textwidth}
        \centering
    \includegraphics[width=1.0\linewidth]{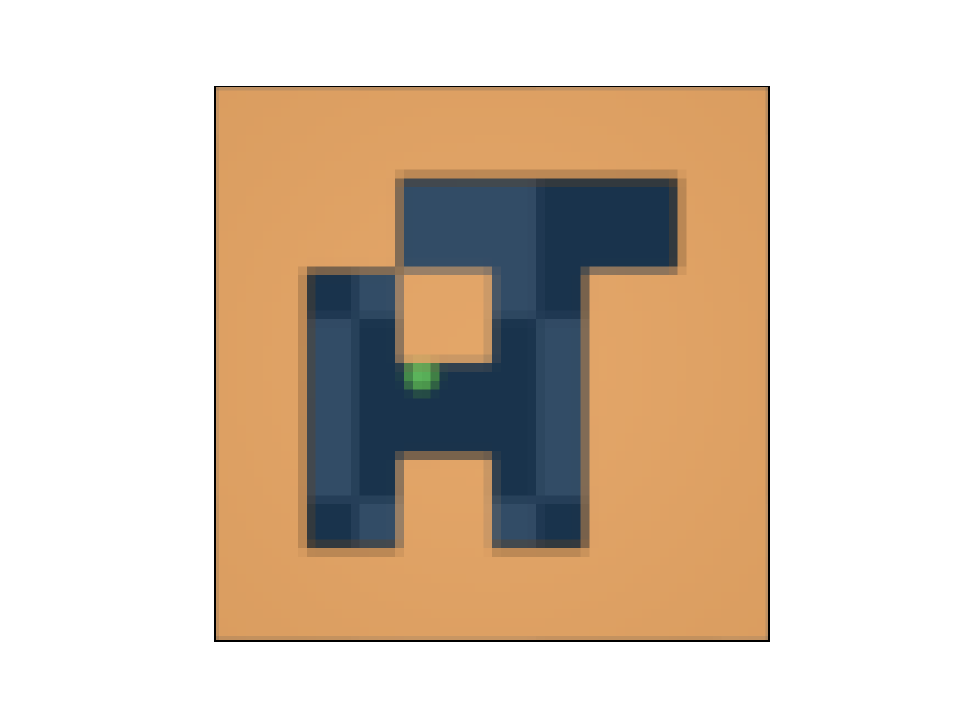} 
    \end{subfigure}
        ~
    \begin{subfigure}[b]{0.18\textwidth}
        \centering
    \includegraphics[width=1.0\linewidth]{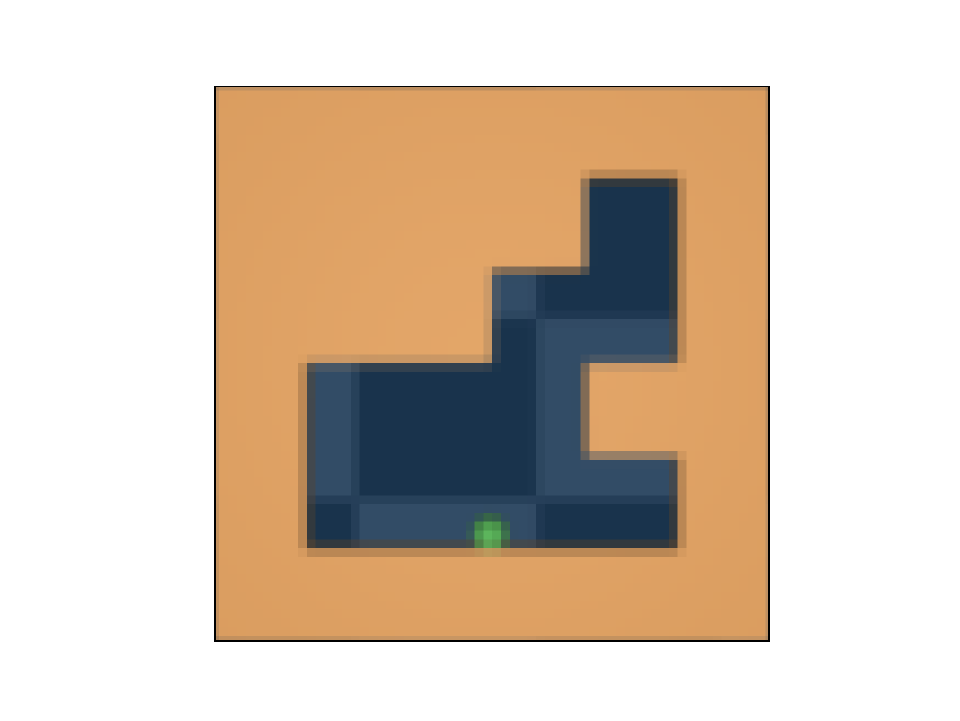} 
    \end{subfigure}
        ~
    \begin{subfigure}[b]{0.18\textwidth}
        \centering
    \includegraphics[width=1.0\linewidth]{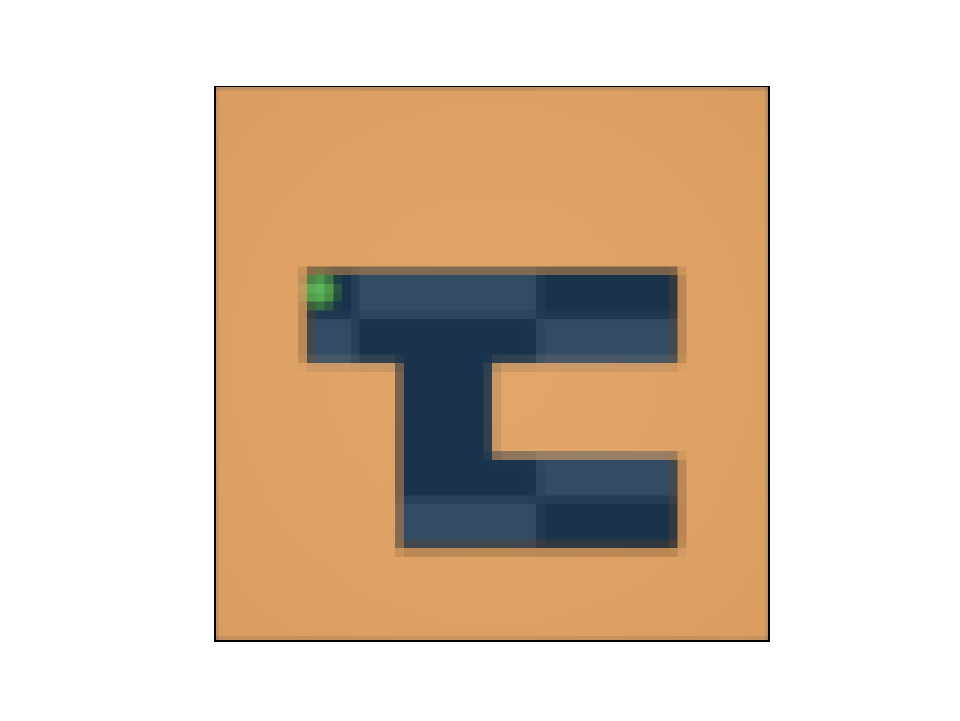} 
    \end{subfigure}
        ~
    \begin{subfigure}[b]{0.18\textwidth}
        \centering
    \includegraphics[width=1.0\linewidth]{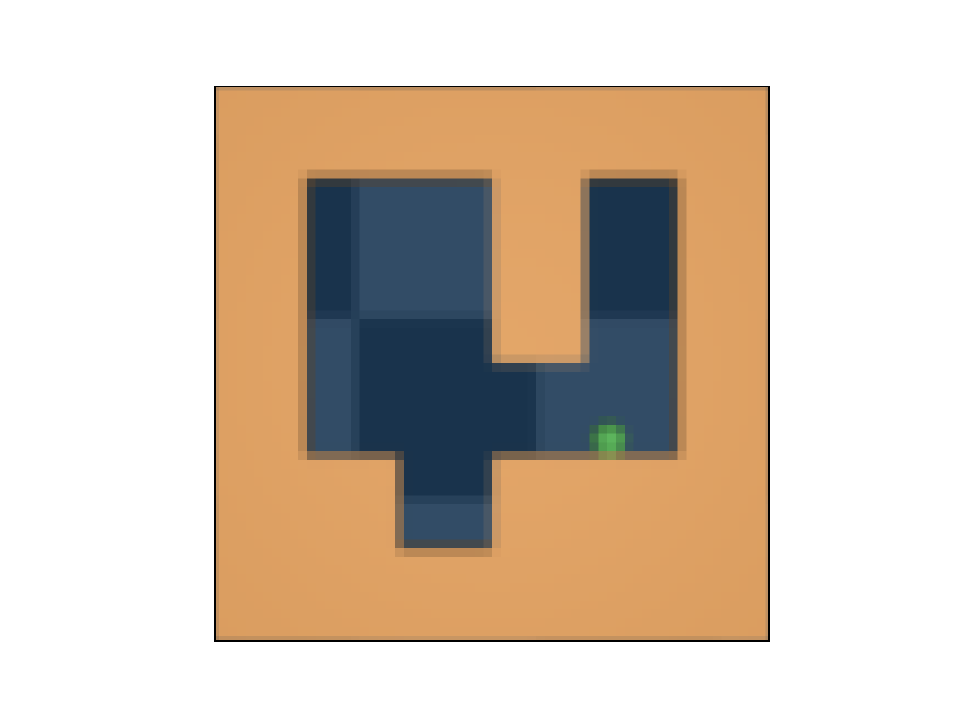} 
    \end{subfigure}
    \caption{Examples of states of the maze environment (the agent is the green point)}
    \label{fig:mazetrain}
\end{figure}

\subsection{Experiment settings}
\label{Expsettings}

The code used for the experiments is based on an implementation of JEPA models for action planning by \cite{sobal2025learningrewardfreeofflinedata}.

In the models, we use flat representations of size 512, a predictor with a MLP architecture and an action encoder set to the identity. The state encoder is based on a simple architecture combining convolutions and residual connections. Before being passed to the predictor, the representations of states and actions are concatenated. The encoder has 2.2M parameters and the predictor has 1.3M parameters. The input trajectories are subsampled into segments of length 16 during training.

All networks were trained with a base learning rate of 0.0028, using the Adam optimizer and a cosine learning rate schedule. For the wall environments, the VCReg loss is computed along the batch dimension of the representations. At planning time, the MPPI optimization in the MPC is configured with 2000 initial perturbations sampled from a Gaussian distribution with mean 0 and standard deviation 12, and a temperature parameter of $\lambda = 0.005$. We use a planning horizon of 96 for a total of 200 planning steps in the WS environment, and a planning horizon of 64 for a total of 64 planning steps in the WB environment. For the maze environment, the VCReg loss is computed along both the batch and temporal dimensions. The MPPI optimization in the MPC is configured with 500 initial perturbations sampled from a Gaussian distribution with mean 0 and standard deviation 5, and a temperature parameter of $\lambda = 0.0025$. We use a planning horizon of 100 for a total of 200 planning steps.

For all experiments, we use $\gamma = 0.98$ and $\tau = 0.80$ for the VF-based approaches, and $\gamma = 0.93$ and $\tau = 0.60$ for the VF\_quasi-based ones. These values were optimized following the procedure described in Appendix \ref{Addexp}

\subsection{Additional experiments}
\label{Addexp}

\textbf{Hyperparameter optimization.} Before running our experiments, we optimized the two main hyperparameters controlling the behavior of the value function learning methods, namely $\tau$ and $\gamma$. This was done using a WS dataset different from the one used for the rest of the tests. The results are shown below:

 \begin{figure}[h]
     \centering
     \begin{subfigure}[b]{0.45\textwidth}
         \centering
     \includegraphics[width=1.0\linewidth]{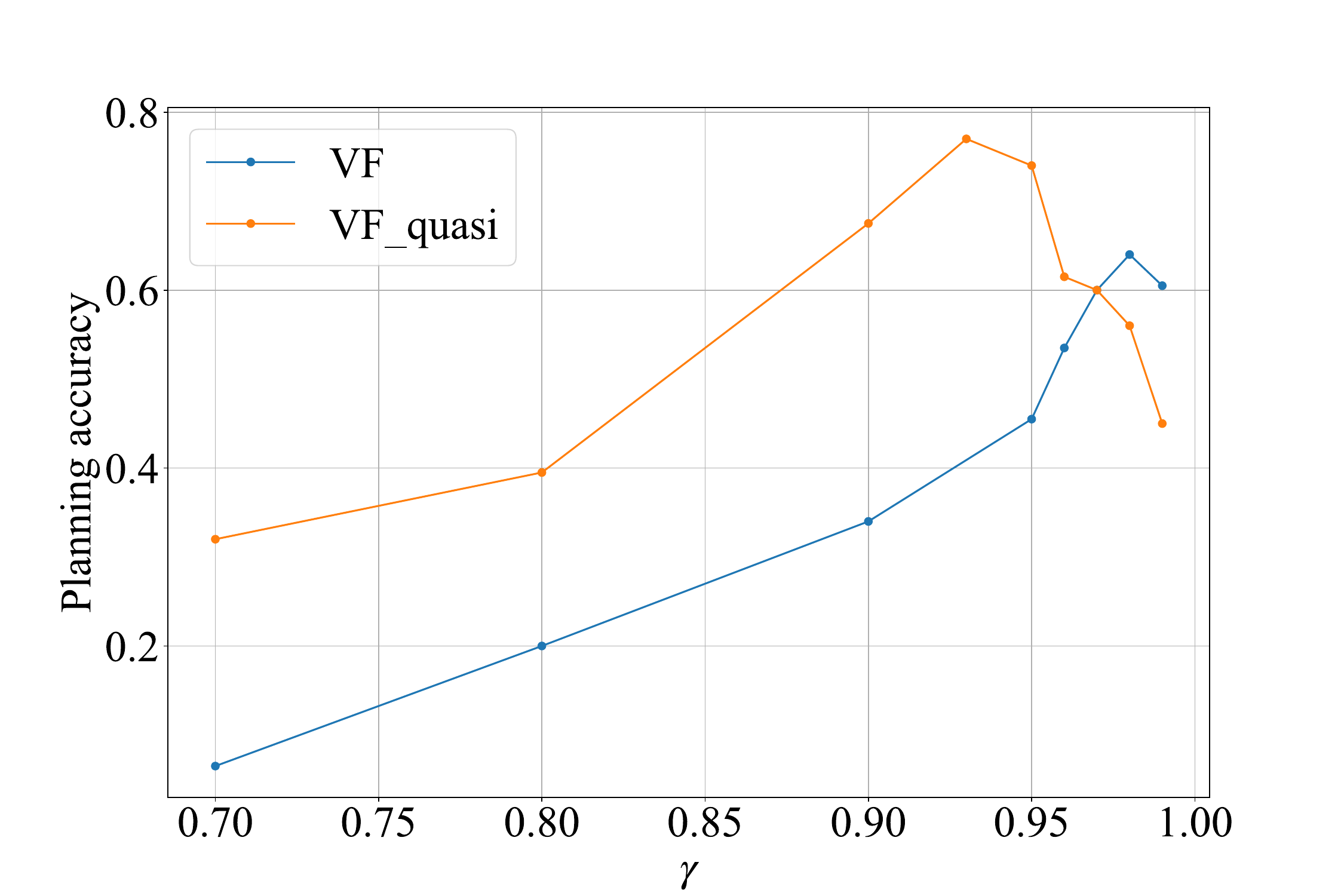} 
     \caption*{Results for $\gamma$ ($\tau=0.7$)}
     \end{subfigure}%
     ~
     \begin{subfigure}[b]{0.5\textwidth}
         \centering
    \includegraphics[width=1.0\linewidth]{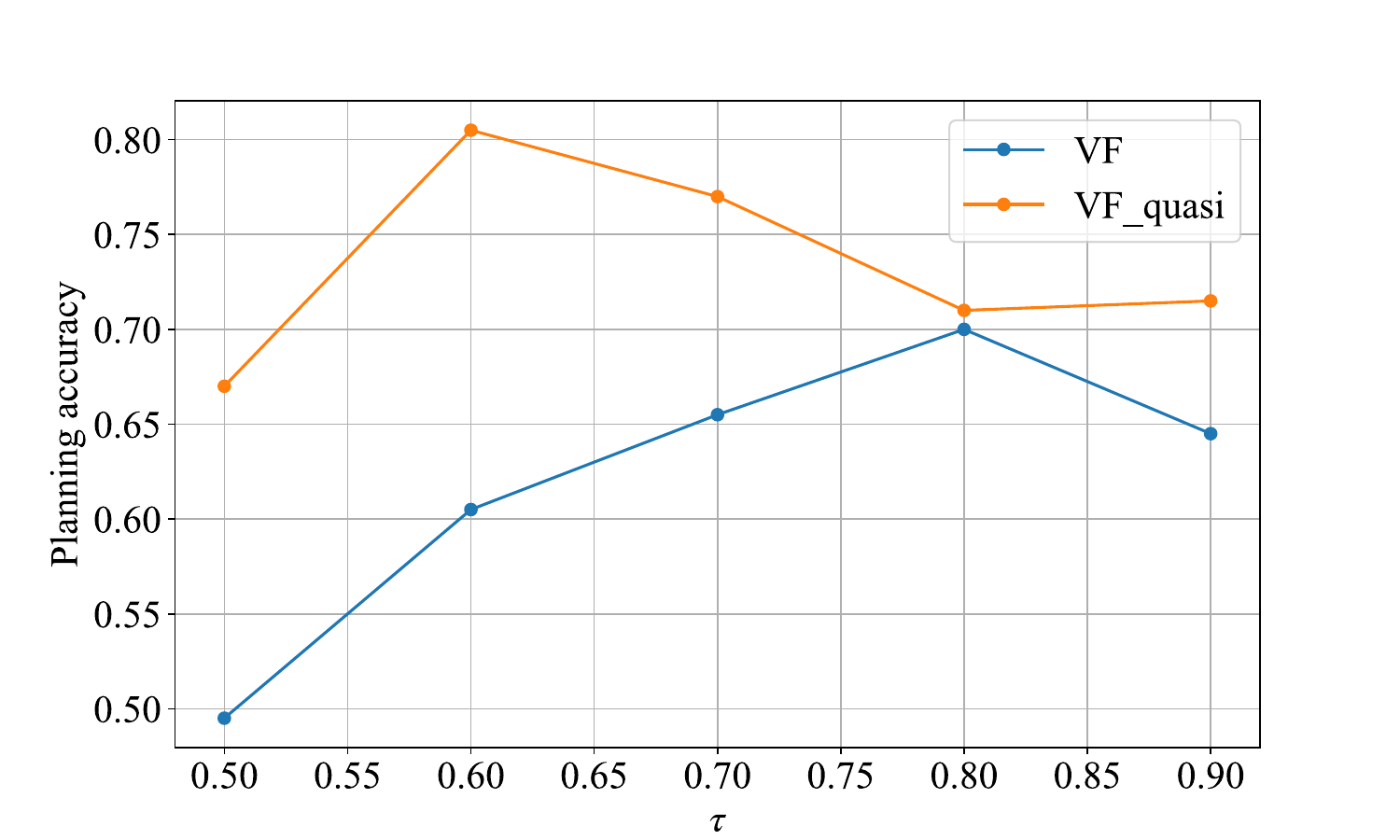} 
     \caption*{Results for $\tau$ (VF: $\gamma=0.98$, VF\_quasi: $\gamma=0.93$)}
     \end{subfigure}
     \caption{Evolution of planning accuracy with respect to hyperparameters}
     \label{fig:}
 \end{figure}

Increasing $\gamma$ improves performance, as it better captures the relationships between distant states. The same applies to $\tau$, which should theoretically be set as close to $1$ as possible. However, setting either parameter too close to $1$ introduces instabilities that degrade performance. We chose the values of $\gamma$ and $\tau$ that maximized the planning accuracy for the rest of the experiments.

\textbf{Separate predictive and planning representations.} One might hypothesize that representations learned using a prediction loss yield better prediction accuracy, while those learned with an IQL approach result in a more effective planning cost. It is possible to combine the advantages of both by adopting an intermediate approach. In this approach, two separate representation spaces are learned: the first with a standard prediction loss, and the second with an IQL loss using a second state encoder. During planning, the first level is used to compute predictions, and the second level to compute the cost. We tested this method with the WS dataset using the pred\_VCReg approach for the first level and the VF approach for the second level. It did not improve planning results, with a planning accuracy of $0.60$.

\end{document}